\documentclass[runningheads]{llncs}

\usepackage[T1]{fontenc}
\usepackage{graphicx}
\usepackage{booktabs}
\usepackage[misc]{ifsym}

\usepackage[ruled,vlined]{algorithm2e}
\usepackage{amsmath, amssymb}
\usepackage{orcidlink}
\usepackage{mwe}
\usepackage{booktabs,multirow,array,graphicx}
\usepackage[table]{xcolor}
\usepackage[normalem]{ulem} 
\usepackage{booktabs,multirow,array,graphicx}
\usepackage[dvipsnames]{xcolor}
\usepackage[normalem]{ulem} 
\usepackage{adjustbox}      
\usepackage{subcaption}
\usepackage{amsmath}
\usepackage[spaces,hyphens]{xurl}
\usepackage{booktabs}
\usepackage{multirow}
\usepackage[table]{xcolor}

\usepackage{graphicx}
\usepackage{subcaption}
\usepackage{tikz}
\usepackage{wrapfig}

\usepackage[margin=1.715in]{geometry}
\usepackage[spaces,hyphens]{xurl}
\usepackage{amssymb}
\usepackage{textcomp}

\newcommand{\fref}[1]{Fig.~\ref{#1}}

\newcommand{\sref}[1]{Section~\ref{#1}}

\usepackage{hyperref}
\usepackage{enumitem} 
\title{{Learning What Can Be Picked: Active Reachability Estimation for Efficient Robotic Fruit  Harvesting}}

\author{
Nur Afsa Syeda \and %
Mohamed Elmahallawy\orcidlink{0000-0002-5731-9253}\thanks{Corresponding author.  \newline To appear in Proceedings of PAKDD 2026: {\em The} $30^{th}$ {\em Pacific-Asia Conference on Knowledge Discovery and Data Mining.}}\and
John Miller\and
Luis de la Torre 
}
\authorrunning{Nur Afsa Syeda and Mohamed Elmahallawy}
\institute{
Washington State University, Richland, WA 99354, USA \\
\email{\{nurafsa.syeda, mohamed.elmahallawy, 
	jmiller16,luis.delatorre\}@wsu.edu}
}

\begin{document}
\maketitle

\vspace{-0.5cm}

\begin{abstract}

Agriculture remains a cornerstone of global health and economic sustainability, yet labor-intensive tasks such as harvesting high-value crops continue to face growing workforce shortages. Robotic harvesting systems offer a promising solution; however, their deployment in unstructured orchard environments is constrained by inefficient perception-to-action pipelines. In particular, existing approaches often rely on exhaustive inverse kinematics or motion planning to determine whether a target fruit is reachable, leading to unnecessary computation and delayed decision-making. Our approach combines RGB-D perception with active learning to directly learn reachability as a binary decision problem.  We then leverage active learning to selectively query the most informative samples for reachability labeling, significantly reducing annotation effort while maintaining high predictive accuracy. Extensive experiments demonstrate that the proposed framework achieves accurate reachability prediction with \emph{substantially fewer labeled samples}, yielding approximately 6–8\% higher accuracy than random sampling and enabling label-efficient adaptation to new orchard configurations. Among the evaluated strategies, entropy- and margin-based sampling outperform Query-by-Committee and standard uncertainty sampling in low-label regimes, while all strategies converge to comparable performance as the labeled set grows. These results highlight the effectiveness of active learning for task-level perception in agricultural robotics and position our approach as a scalable alternative to computation-heavy kinematic reachability analysis. Our code is available through \url{https://github.com/wsu-cyber-security-lab-ai/active-learning}.

\vspace{-0.3cm}
\keywords{Apple harvesting system, Active Learning, Computer Vision, Image processing, Classification}\vspace{-0.2cm}
\end{abstract}

\vspace{-0.9cm}
\section{Introduction}\label{sec:intro}
\vspace{-0.2cm}
Agriculture is among the oldest and most vital economic activities sustaining human society, with fruit harvesting playing a central role in global food security and economic stability \cite{pawlak2020role}. Among fruit crops, apples have attracted significant attention due to their high market value, widespread consumption, and labor-intensive harvesting requirements. USA is one of the world’s largest apple producers, with Washington State alone accounting for approximately 68\% of national production in 2024 \cite{USDA_NASS_FR08_2024}. 
However, this success increasingly comes at the cost of rising production expenses, climate variability, environmental constraints, and, most critically, persistent labor shortages \cite{win}. Specifically, labor availability has emerged as a dominant bottleneck in apple harvesting, which still relies heavily on manual hand-picking to preserve fruit quality. This process is time-consuming, physically demanding, and requires a large, seasonal workforce. The challenge is exacerbated by demographic shifts and increasingly restrictive immigration policies, which have significantly reduced the availability of skilled farm labor. As a result, apple producers face mounting pressure to adopt scalable, automated harvesting solutions that can operate reliably in unstructured orchard environments \cite{SimnittApplePicking,agriengineering5040136}.


Robotic harvesting systems, enabled by advances in artificial intelligence, perception, and manipulation, offer a promising pathway toward addressing these challenges. Early agricultural robots focused primarily on autonomous navigation, while subsequent generations incorporated machine vision and robotic manipulators for tasks such as fruit detection, harvesting, pruning, and monitoring \cite{7238364,NAVONE2025101283}. Notably, existing research has played a pivotal role in advancing mechanized apple harvesting systems, establishing foundational contributions in robotic end-effector design, motion analysis, and real-world field deployment \cite{b1}. Despite this progress, deploying robotic harvesters at scale remains difficult due to the complex interaction between perception, decision-making, and manipulation under real-world constraints.

A key yet underexplored challenge lies in ``reachability reasoning''--determining whether a detected fruit can be physically reached and harvested by a robotic arm before invoking computationally expensive motion planning or inverse kinematics (IK). Existing systems often attempt to solve full kinematic feasibility for each detected apple, resulting in unnecessary computation, delayed decisions, and reduced throughput, particularly in cluttered canopies with partial occlusions and variable lighting \cite{b3,b5}. Moreover, acquiring large, fully labeled datasets that capture reachability across diverse orchard configurations is costly and impractical, limiting the adaptability of learning-based solutions \cite{agriculture13030540}.


In this work, we address these challenges by framing reachability estimation as a task-level decision problem rather than a byproduct of low-level kinematic planning. Specifically, we investigate two fundamental research questions: (i) can fruits be reliably detected and localized from RGB-D imagery in dense orchard environments, and (ii) can reachability be learned efficiently with limited labeled data? To this end, we employ a YOLO-based object detection model to identify fruit (apple) candidates from aligned RGB-D images and propose an active learning–driven framework that selectively queries the most informative samples to learn reachability. 
In summary, our key contributions are:

\begin{itemize}[leftmargin=*]
    \item We propose a \emph{learning-based reachability estimation framework} that enables robotic harvesters to determine fruit accessibility \emph{prior to} invoking joint-level motion planning or inverse kinematics.
    
    \item We integrate \emph{RGB-D perception} with \emph{active learning} to achieve label-efficient reachability prediction in unstructured orchard environments, significantly reducing annotation costs.
    
    \item We systematically evaluate multiple active learning strategies and demonstrate that \emph{entropy- and margin-based sampling} consistently outperform uncertainty sampling and Query-by-Committee in low- and moderate-label regimes.
    
    \item Through extensive experimental validation, we show that \emph{decision-level reachability reasoning} reduces the number of inverse-kinematics feasibility checks required by filtering unreachable targets before planning, thereby eliminating IK evaluation for approximately 38\% of detected candidates per frame. The Active learning improves reachability prediction accuracy by approximately 6-8\% over random sampling and 2-4\% over QBC and uncertainty sampling in low-label settings, positioning our approach as a scalable alternative to exhaustive kinematic reachability analysis. 
\end{itemize}

\vspace{-0.6cm}

\section{Related Work}
\label{sec:rel_work}
\vspace{-0.2cm}

Significant prior work has advanced automation in agriculture by combining perception, learning, and robotic manipulation to reduce the reliance on manual labor. Early and recent efforts have produced a variety of systems for seeding, fruit detection, grasping, and end-effector design, demonstrating both the promise and the remaining challenges for large-scale deployment \cite{10825241,zhang2024automated,navas2024soft,yin2025design,li2023design,b11,coll2023accurate}. 

\noindent{\bf \em Perception for Fruit Detection.}
A substantial line of research focuses on visual detection and segmentation of fruit in challenging orchard scenes. Kuznetsova et al. \cite{b13} demonstrated a real-time apple detection pipeline based on YOLOv3 with tailored pre- and post-processing steps to handle occlusions and overlapping fruit; their system reported detection times on the order of tens of milliseconds per instance.  Similarly, dataset and RGB-D acquisition efforts (e.g., Zhang et al. \cite{b4}) have enabled more robust learning and evaluation in full-foliage conditions.  

\noindent{\bf \em Robotic Harvesting and Manipulation.}
Research on robotic harvesting spans end-effector design, kinematic and dynamic analysis, and simulation-aided system evaluation.  Prior studies have explored semi-automated shaking prototypes, soft end-effectors, and full-system field trials that analyze the physical interaction between manipulators and fruit \cite{zhang2024automated,navas2024soft,yin2025design}. Simulation and motion analysis have informed mechanical design and planning heuristics~\cite{li2023design}. 


\noindent{\bf \em Active learning and Label Efficiency.}
Label scarcity is a pervasive issue for supervised methods in agricultural settings. Active learning strategies, surveyed comprehensively by Settles \cite{b6}, selectively query labels to reduce annotation effort while preserving model performance. Foundational active learning methods include uncertainty based sampling, Query-by-Committee (QBC) \cite{b10}, and sequential selection algorithms for text and vision tasks. Recent applied work confirms its utility in domain-specific annotation-scarce settings \cite{b14,b15}. More recent studies further explore label-efficient learning in practical and multimodal settings, demonstrating improved annotation efficiency and robustness in real-world scenarios \cite{10825322,shreen}.


\begin{figure*}[!t]    
    \includegraphics[width=\linewidth]{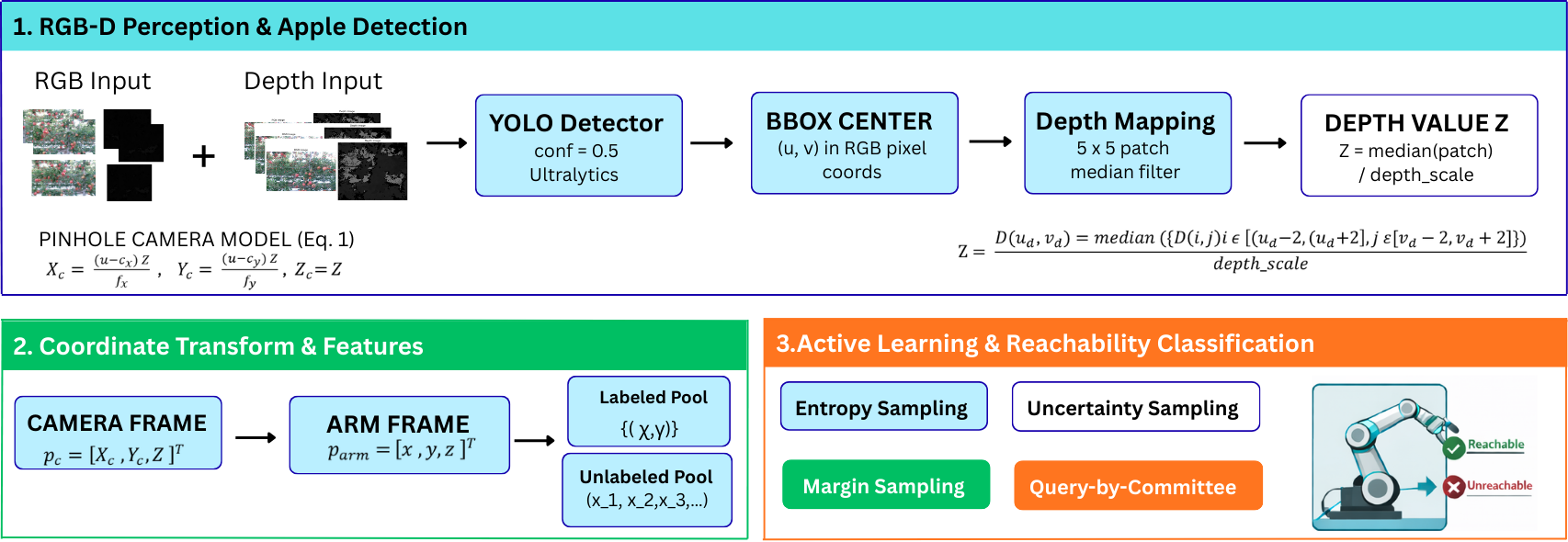} 
    \caption{Overview of the proposed reachability estimation framework.} 
     \label{framework_overall}    \vspace{-5mm}
\end{figure*}

\noindent{\bf \em Gaps and Motivation for Decision-Level Reachability.}
While prior work provides strong solutions for fruit detection and for physical harvesting mechanisms, relatively little research addresses \emph{task-level} reachability: a fast, reliable decision that answers whether a detected fruit is worth attempting (i.e., reachable) before invoking computationally expensive planners or executing risky arm motions. Existing pipelines typically rely on per-candidate IK or full motion planning, which are costly and fragile in cluttered, partially observed canopies. Moreover, few works combine RGB-D perception with label-efficient learning strategies specifically targeted at reachability prediction.

To address this gap, we integrate RGB-D perception with active learning to build a decision-level reachability predictor that (i) reduces reliance on exhaustive kinematic checks, (ii) lowers labeling cost, and (iii) remains robust under orchard occlusion and variability. Recent robotics foundation models, such as RT-2~\cite{zitkovich2023rt} and OpenVLA~\cite{kim2024openvla}, suggest that such capabilities may eventually be embedded in larger policies. We view them as complementary: unlike these resource-intensive models, our method operates on a compact nine-dimensional feature vector with fewer than 1,000 labeled samples. The next section presents our YOLO-based pipeline, evaluated active learning strategies, and their integration into a scalable reachability module.


\vspace{-1mm}
\section{Methodology}
\label{sec:method}
\vspace{-2mm}
This section describes our end-to-end methodology for learning decision-level reachability in a robotic apple-harvesting system.  
\fref{framework_overall} provides an illustration the overall framework of the proposed methodology, from RGB-D perception and 3D localization to active learning–based dataset construction and reachability classification.


\begin{itemize}[leftmargin=*] 
    \item \textbf{RGB-D acquisition:} Capture synchronized RGB and depth images from orchard environments.  
    \item \textbf{Detection:} Identify candidate apples in RGB images using a YOLO-based deep learning detector.  
    \item \textbf{Depth mapping:} Map detected apple pixels to depth information, aggregating local patches to reduce noise and handle missing values.  
    \item \textbf{Coordinate transformation:} Transform 3D candidate points from camera coordinates to the manipulator frame for actionable robot input.  
    \item \textbf{Dataset preparation:} Compile labeled feature vectors combining RGB, depth, and geometric features for training reachability models.  
\item \textbf{Classifier:} Map extracted features to reachability predictions using a Random Forest classifier mapping feature vector $\phi(p_{\text{arm}})$ to a binary reachability prediction.
    
    \item \textbf{Active learning:} Iteratively select the most informative samples from the unlabeled pool to maximize labeling efficiency and improve model performance.  
\end{itemize}
Together, these components form a theoretical pipeline that tightly couples perception, geometric reasoning, and active learning, enabling a fast, accurate, and label-efficient reachability predictor for robotic manipulation in orchard environments.

\begin{wrapfigure}{r}{0.6\linewidth}  
    \centering \vspace{-17mm}
    \includegraphics[width=0.49\linewidth]{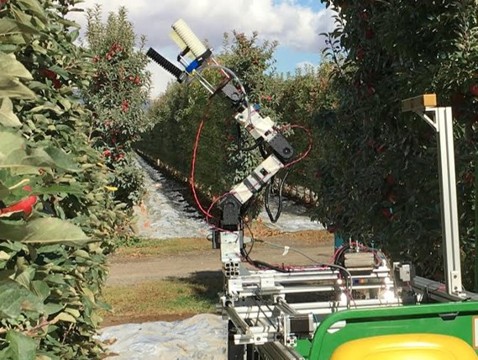}
    \includegraphics[width=0.49\linewidth]{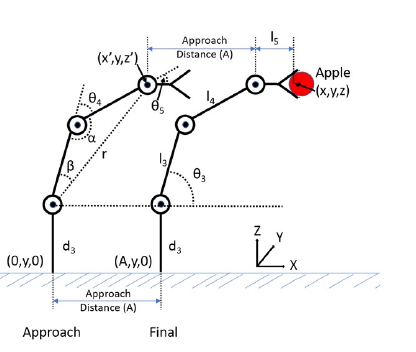}
    \caption{Robotic apple-harvesting system (left) and corresponding 2D picking-sequence visualization (right)~\cite{b3}.}\vspace{-25pt}
    \label{arm_apple}
\end{wrapfigure}

\vspace{-2mm}
\subsection{System Overview}
\vspace{-1mm}
The field platform (\fref{arm_apple}) is a 5-DOF manipulator composed of two prismatic joints (cartesian motion along the orchard aisle and toward/away from the fruit wall) and three revolute joints. The end-effector is held at a fixed pitch (horizontal) during the picking sequence and is designed to grip and pull fruit while minimizing collisions with branches and neighboring fruit.

\begin{figure*}[!t]
  \centering
  \begin{subfigure}[t]{0.32\textwidth}
    \centering
    \includegraphics[width=\linewidth]{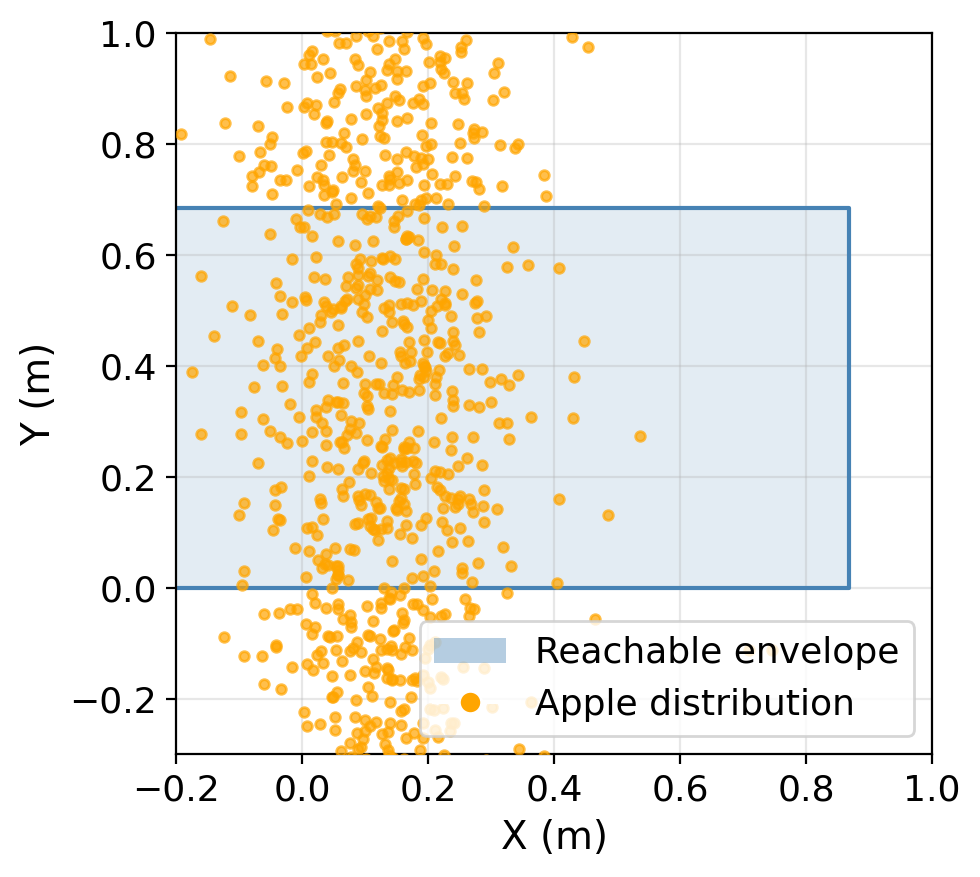}
    \caption{Top view.}
  \end{subfigure}
  \hfill
  \begin{subfigure}[t]{0.32\textwidth}
    \centering
    \includegraphics[width=\linewidth]{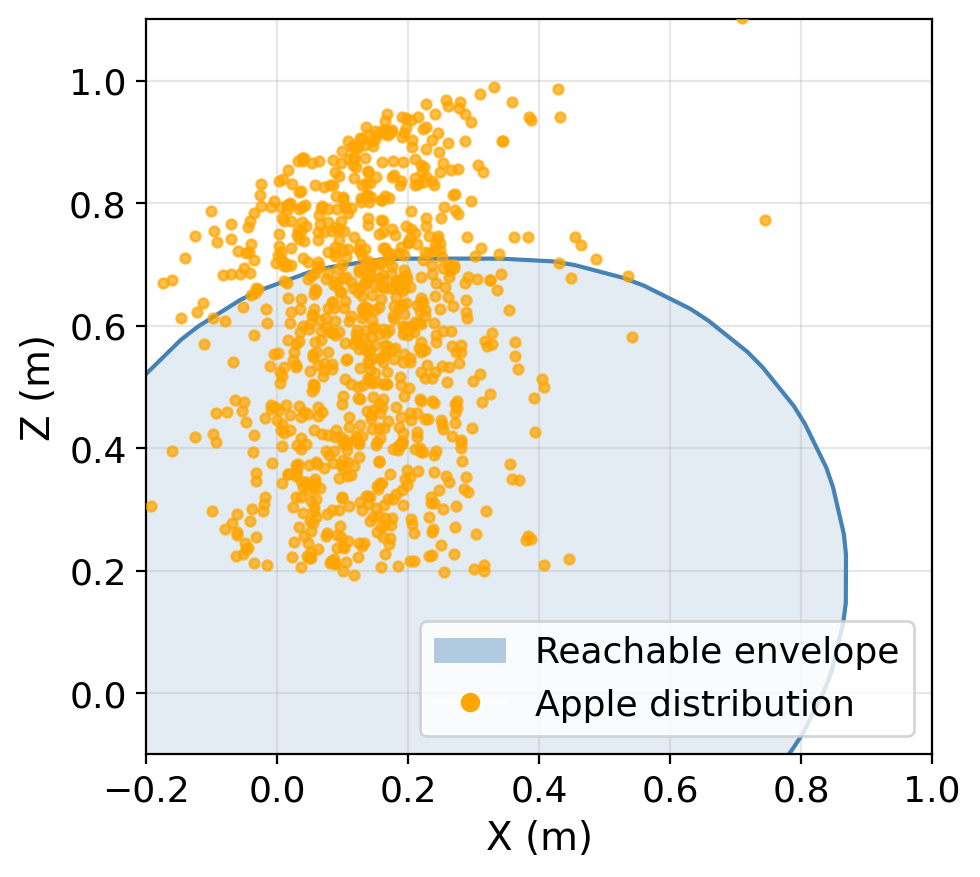}
    \caption{Side view.}
  \end{subfigure}
  \hfill
  \begin{subfigure}[t]{0.32\textwidth}
    \centering
    \includegraphics[width=\linewidth]{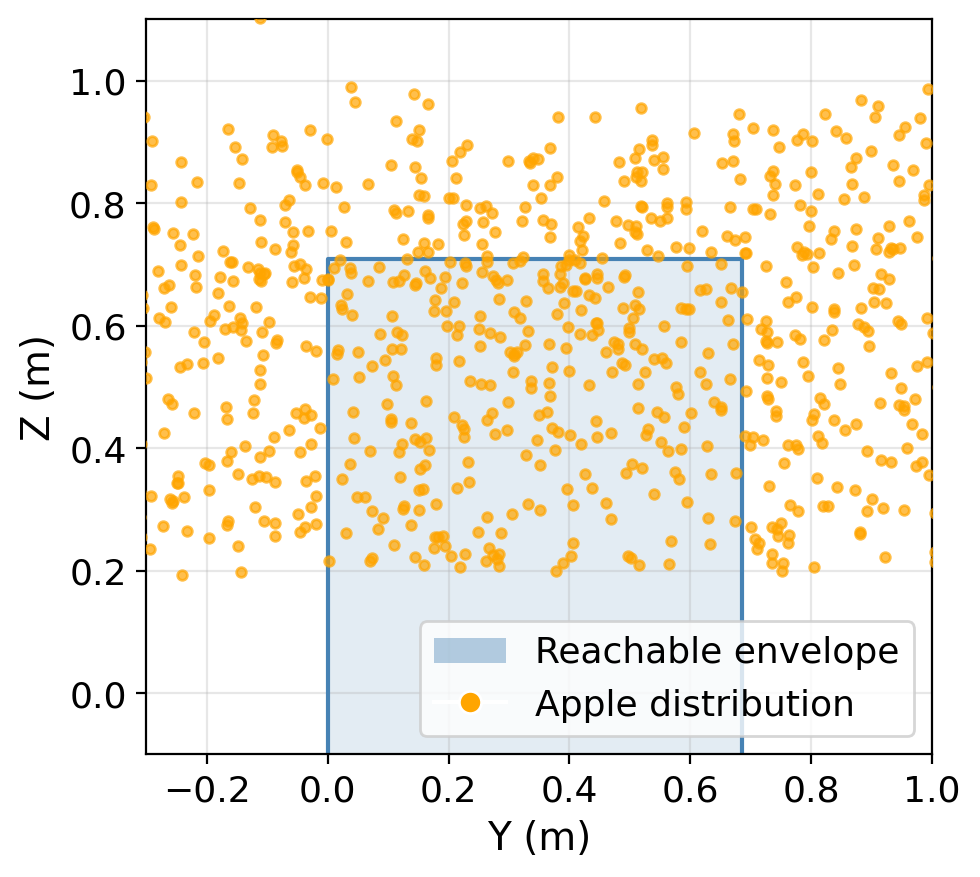}
    \caption{Front view.}
  \end{subfigure}
\caption{Reachable envelope of the manipulator (blue shaded region) overlaid with apple coordinate distribution from field data in arm-frame coordinates (orange points).}\label{workspace}
\end{figure*}

\fref{workspace} illustrates the reachable envelope of the manipulator, derived by forward kinematics simulation over all joint variable combinations within their physical limits, overlaid with apple coordinates extracted from field data in the arm coordinate frame. 
As shown in \fref{workspace}, a substantial portion of detected apples fall outside the reachable envelope. This partial overlap between the reachable workspace and the fruit distribution motivates a decision-level reachability module that can quickly determine whether a detected fruit falls within the manipulator's reach before invoking kinematic evaluation, reducing unnecessary IK calls per frame.


\subsection{Data Collection and Preprocessing}

RGB--depth image pairs were collected in commercial orchards across Washington State, USA, using a Kinect v2 sensor (RGB: $1920 \times 1080$ BMP; depth: $512 \times 424$ PNG). The dataset is organized by date-stamped collection sessions. The dataset comprises 3,480 synchronized RGB--depth pairs, of which a working subset of 2,519 pairs was used for detection and subsequent 3D processing. Data collection spanned multiple days and naturally encompassed varying lighting conditions typical of field harvesting operations, including partial canopy shade and differing sun angles.

\begin{wrapfigure}{r}{0.55\linewidth} 
    \centering
   \vspace{-1mm}
    \includegraphics[width=0.49\linewidth]{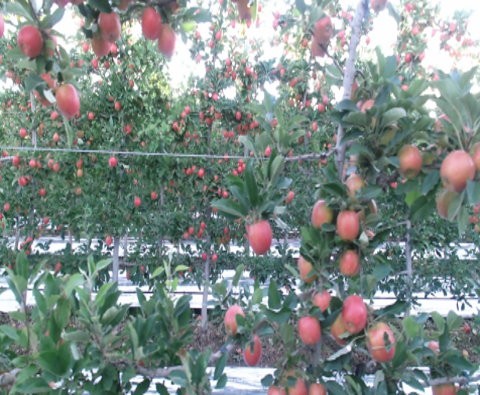}
    \includegraphics[width=0.49\linewidth]{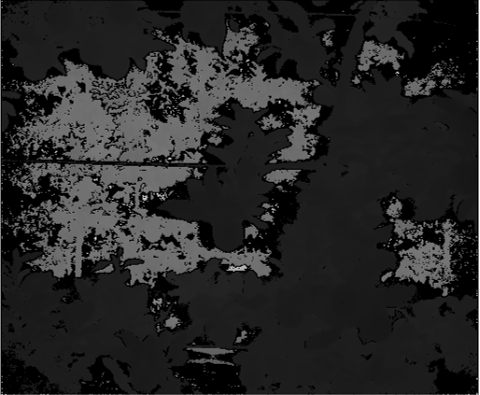}
    \caption{Representative RGB image of an apple (left) and the corresponding depth projection (right) from a full-foliage orchard dataset.}
    \label{fig:rgb_sample_wrap}  \vspace{-8mm}
\end{wrapfigure}
Apple detection is performed by applying a YOLO-based model directly to each RGB image using a fixed confidence threshold of 0.5, without illumination-specific preprocessing.
For each detected apple, we extract a small depth patch ($5 \times 5$ pixels) centered at the corresponding mapped pixel and use the median depth value to suppress sensor noise and outliers. \fref{fig:rgb_sample_wrap} shows a sample RGB image with corresponding depth projections. Detections near image boundaries or with invalid/zero depth values are discarded. 

\vspace{-2mm}
\subsection{Perception and Pixel-to-Point Mapping}
We detect candidate apples using a YOLO-based object detector (Ultralytics pretrained model) applied to the RGB image. For each detection bounding box, we compute the bounding-box center \((u,v)\) in RGB pixel coordinates and map it to the depth image using the appropriate scale factors for resolution differences. Given the measured depth \(Z\) at the mapped pixel, we compute camera-frame coordinates \((X_c,Y_c,Z_c)\) via the pinhole camera model:
\begin{equation}
    X_c = \frac{(u - c_x)\, Z}{f_x}, \qquad
    Y_c = \frac{(v - c_y)\, Z}{f_y}, \qquad
    Z_c = Z,
\end{equation}
where \((c_x,c_y)\) and \((f_x,f_y)\) are the camera principal point and focal lengths (intrinsic parameters).

\vspace{-3mm}
\subsection{Extrinsic Calibration and Transform to Manipulator Frame}
To reason about reachability, we express each candidate point in the manipulator (arm) reference frame. We compute the rigid-body transform from camera to arm frame using a standard extrinsic calibration procedure (checkerboard/hand–eye calibration). Denote the rotation \(R\in \mathbb{R}^{3\times3}\) and translation \(t\in\mathbb{R}^{3}\) from camera to arm frame. Then, the arm-frame coordinates \(p_{\text{arm}}\) are:
\begin{equation}
    p_{\text{arm}} = R\, p_{c} + t,\qquad p_c = [X_c, Y_c, Z_c]^\top.
\end{equation}
In our deployment, the calibration translation (measured during the field study) was \(t=[0.76,\;0.44,\;0.485]^\top\) (meters). If the mounting provided negligible rotation, we approximate the rotation matrix as the $3\times3$ identity matrix, $R \approx I_{3\times 3}$;
otherwise, we use the measured rotation matrix.

\vspace{-3mm}
\subsection{Reachability Labels and Dataset Split}
To bootstrap supervised learning, we generate reachability labels using a deterministic kinematic feasibility check: for each candidate 3D point, we attempt to solve a constrained IK check for the manipulator (respecting joint limits, prismatic travel bounds, and a collision-avoidance margin). If the IK solver finds a feasible approach pose for the end-effector that satisfies these constraints, the point is labeled \texttt{reachable}; otherwise, it is labeled \texttt{unreachable}. From the available data, we construct a labeled dataset of 1,000 samples \((p_{\text{arm}}, y)\), where \(y\in\{0,1\}\). The labeled set is partitioned into an initial labeled pool \(\mathcal{L}\) (80\%) and a held-out test set (20\%). An additional unlabeled pool \(\mathcal{U}\) of approximately 24,000 candidate points was used for active learning queries. 

\vspace{-3mm}
\subsection{Feature Representation and Classifier}
Although the 3D coordinates are the primary input, reachability depends on more than the target position alone. For each candidate, we compute a compact feature vector combining geometric and perception cues:
\[
\phi(p_{\text{arm}}) = \big[\,x,y,z,\; \|p_{\text{arm}}\|,\; \theta_{\text{az}},\; \theta_{\text{el}},\; \sigma_Z,\; A_{\text{bbox}},\; D_{\text{local}}\,\big],
\]
where \(\|p_{\text{arm}}\|\) is the Euclidean range to the point, \(\theta_{\text{az}},\theta_{\text{el}}\) are azimuth/elevation angles relative to the arm base, \(\sigma_Z\) is the depth variance in the 5\(\times\)5 patch (occlusion/noise proxy), \(A_{\text{bbox}}\) is the normalized bounding-box area, and \(D_{\text{local}}\) is a local point-density descriptor computed from nearby depth pixels. These features capture both kinematic distance and local scene complexity and are supplied to a binary classifier. In our experiments (\sref{sec:exper}), we use a Random Forest classifier, though MLPs or gradient-boosted trees can also be used.

\vspace{-3mm}
\subsection{Pool-based Active Learning}
To address label scarcity, we use a pool-based active learning loop. Starting from a small labeled set $\mathcal{L}$ and a large unlabeled pool $\mathcal{U}$, the loop iteratively: (i) Trains the reachability classifier on $\mathcal{L}$;(ii) Ranks the instances in $\mathcal{U}$ according to a query criterion; (iii) Selects the top $b$ instances for labeling; (iv) Adds the newly labeled instances to $\mathcal{L}$; (v) Updates the model with the expanded labeled set.
 

\vspace{-2mm}
\subsection{Query Strategies}\label{sec:methods}
We consider several standard, theory-driven active learning query strategies:\begin{itemize}[leftmargin=*]
  \item \textbf{Random sampling:} A baseline that selects instances uniformly from $\mathcal{U}$.
  \item \textbf{Uncertainty sampling:} It selects instances for which the classifier's predicted probability of the most likely class is lowest.
  \item \textbf{Margin sampling:} 
  Offering greater robustness than plain uncertainty in binary tasks.
  \item \textbf{Entropy sampling:} 
  Taking the full label distribution into account.
  \item \textbf{Query-by-Committee (QBC):} 
  Selects instances maximizing disagreement among an ensemble of classifiers.
\end{itemize}
For the IK labeling oracle, we use the deterministic feasibility check described above. This allows us to obtain ground-truth reachability labels without human annotation and enables large-scale, simulation-style experiments in the field.



\vspace{-3mm}
\subsection {Training and Evaluation Protocol}
We adopt an incremental training strategy in which classifiers are updated after each active learning batch using the labeled set \(\mathcal{L}\). Model performance is assessed using standard classification metrics, including accuracy, precision, recall, F1-score, and the area under the ROC curve (AUC). We also introduce the concept of \emph{labeling efficiency}, which characterizes how rapidly model performance improves relative to the number of labeled samples acquired. The efficiency benefit of the proposed framework lies in the reduction of IK invocations rather than the per-candidate speed of the classifier. The trained classifier filters unreachable candidates prior to kinematic evaluation, reducing the number of IK calls required per frame. 






\vspace{-3mm}
\section{Performance Evaluation}\label{sec:exper}

\vspace{-2mm}
\subsection{Experimental Setup}
\noindent{\bf Implementation details.} The pipeline is implemented in Python using the Ultralytics YOLO library for detection and a Random Forest classifier for reachability classification, using the scikit-learn default configuration withhin the modAL active learning framework. Depth information is processed by applying a median filter on 5\(\times\)5 patches and discarding invalid values. Active learning is performed using \texttt{modAL}, with an initial labeled seed of \(|\mathcal{L}|=800\), an unlabeled pool of \(|\mathcal{U}|\approx 24{,}000\), and batch size \(b=50\). 

This implementation realizes the components of a novel pipeline that (i) tightly couples RGB-D perception with geometric feature extraction, (ii) transforms perception outputs into the manipulator frame through principled calibration, and (iii) leverages label-efficient active learning to yield a fast, accurate reachability predictor for cluttered orchard deployment.

\vspace{2mm}
\noindent{\bf Implementation Procedure.} For each configuration of initial labeled size (10, 30, 50) and query budget (50, 100), the classifier was trained incrementally and evaluated on the held-out test set after each batch.



\vspace{2mm}
\noindent{\bf Evaluation metrics.} We evaluate our approach on a held-out test set using standard classification metrics:\text{Accuracy}, \text{Precision},\text{Recall}, \text{F1-score}. Additionally, we report the area under the ROC curve (AUC) and \emph{labeling efficiency} curves, which plot test accuracy versus the number of labeled samples acquired.


\vspace{2mm}
\noindent{\bf Computational Efficiency.} To quantify the reduction in IK invocations achieved by the proposed classifier, we evaluated the fraction of targets filtered as unreachable across the test set. The trained classifier (entropy sampling, $(n_{\text{initial}} = 50,\, n_{\text{queries}} = 100)$) classified 38\% of test candidates as unreachable, eliminating the need for IK evaluation on those candidates entirely. The proposed classifier thereby reduces the total number of kinematic evaluations required per frame, which grows significantly with system complexity and planning overhead.

\subsection{Results and Discussion}

\begin{wrapfigure}{r}{0.5\linewidth}  
    \centering 
 \vspace{-31pt}
    \includegraphics[width=\linewidth]{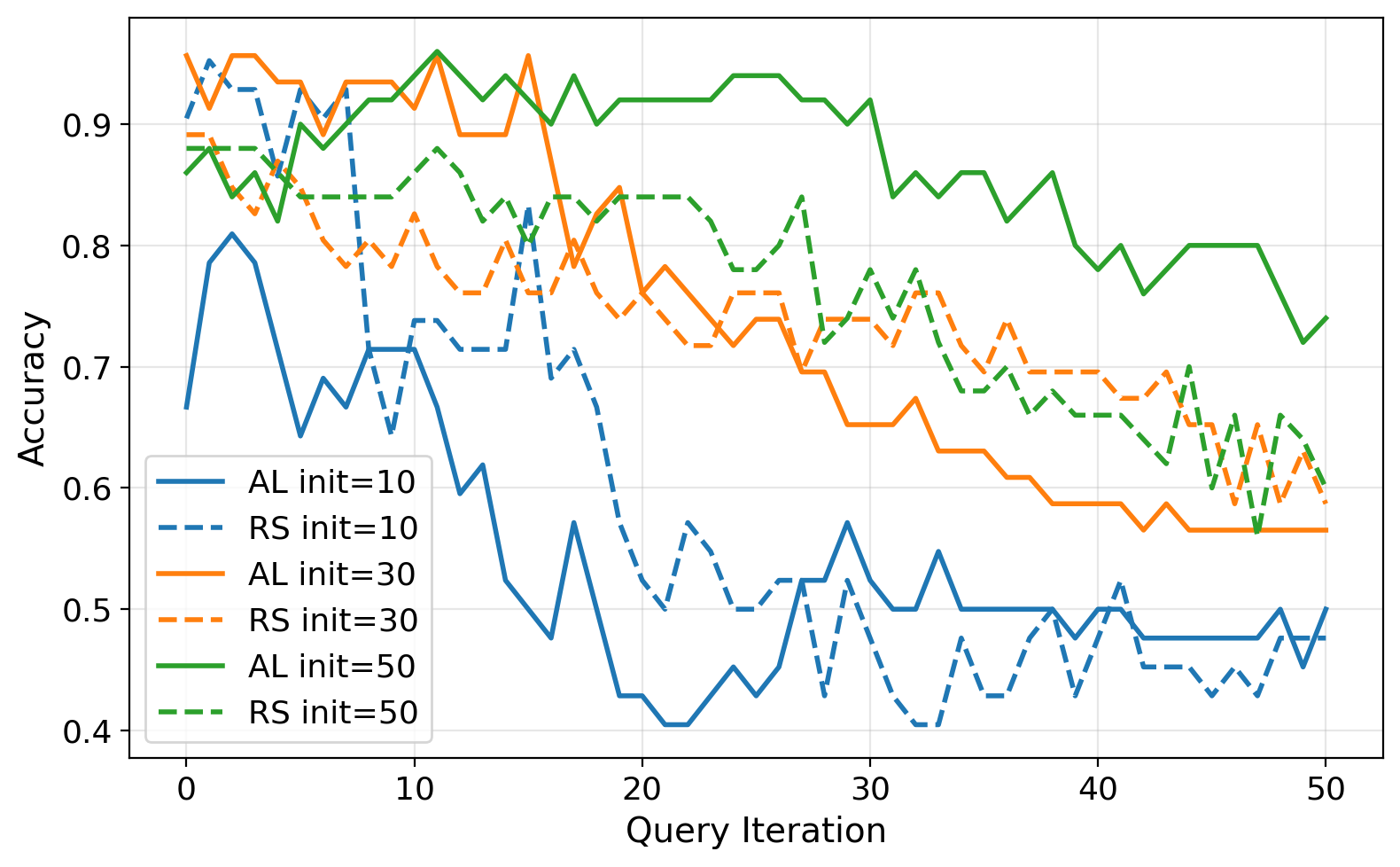}
 
    \includegraphics[width=\linewidth]{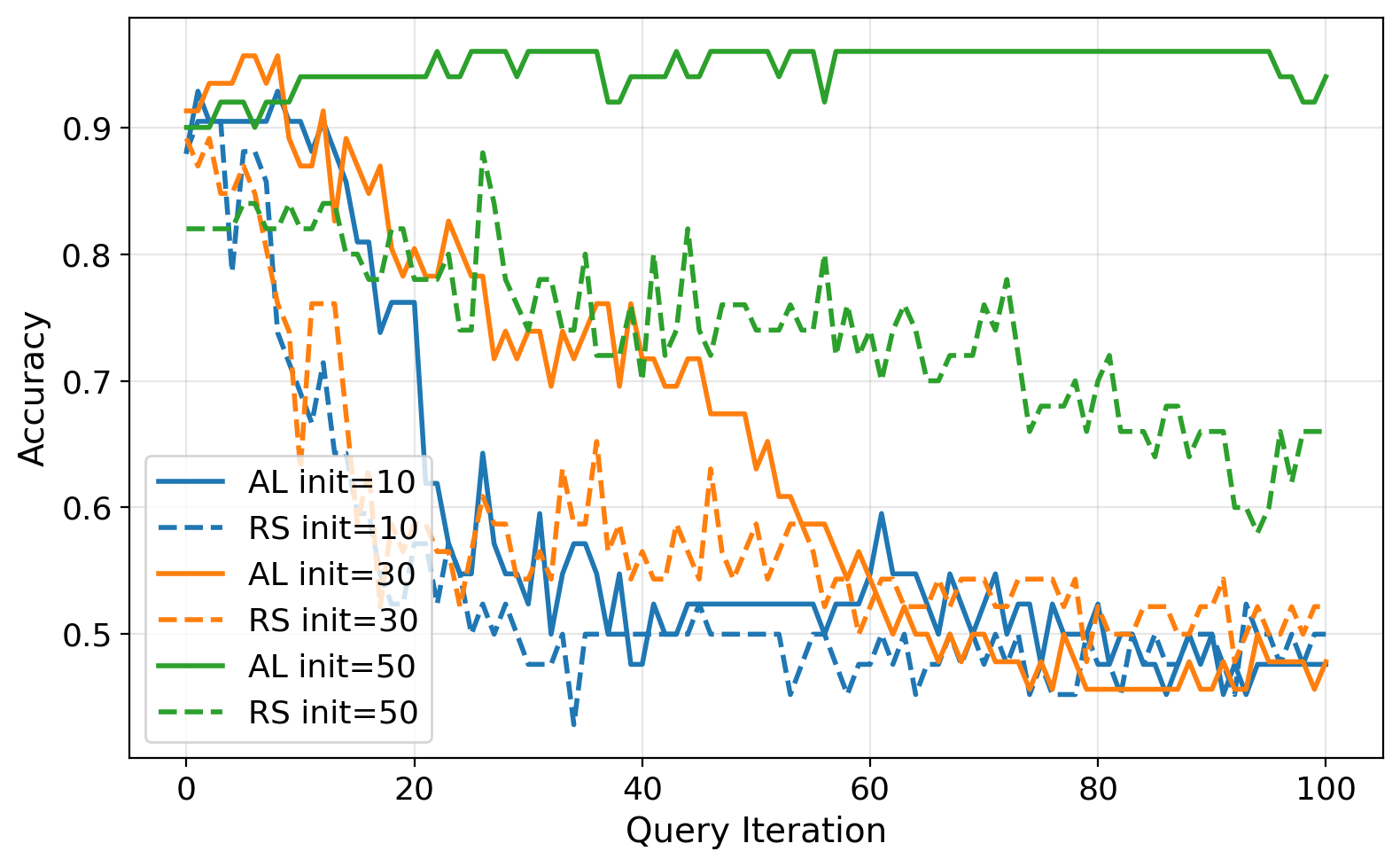}

    \caption{Comparison of learning curves for uncertainty sampling active learning over 50 queries (top) and 100 queries (bottom).}
    \label{fig:single_combined} \vspace{-21pt}
\end{wrapfigure}

\noindent{\bf Comparison of Active Learning and Random Sampling.}\noindent\fref{fig:single_combined} compares the learning curves of active learning (uncertainty sampling) and random sampling for 50 and 100 queries, respectively. Solid lines represent active learning, while dashed lines represent random sampling.

\noindent With a small initial labeled set (size 10), both methods show low and unstable accuracy. As the initial labeled set increases to 30 or 50, active learning generally outperforms random sampling, with the advantage most pronounced in early query iterations. With 100 queries, active learning maintains higher accuracy across all initial label sizes, especially in early stages.

The occasional accuracy dip with more queries is attributable to poorly calibrated probability estimates at small labeled set sizes, which can cause uncertainty-based strategies to temporarily select noisy boundary samples.



\begin{figure*}[!t]
  \centering
  \begin{subfigure}[t]{0.32\textwidth}
    \centering
  \includegraphics[width=\linewidth, trim=0 0 0 0.7cm, clip]{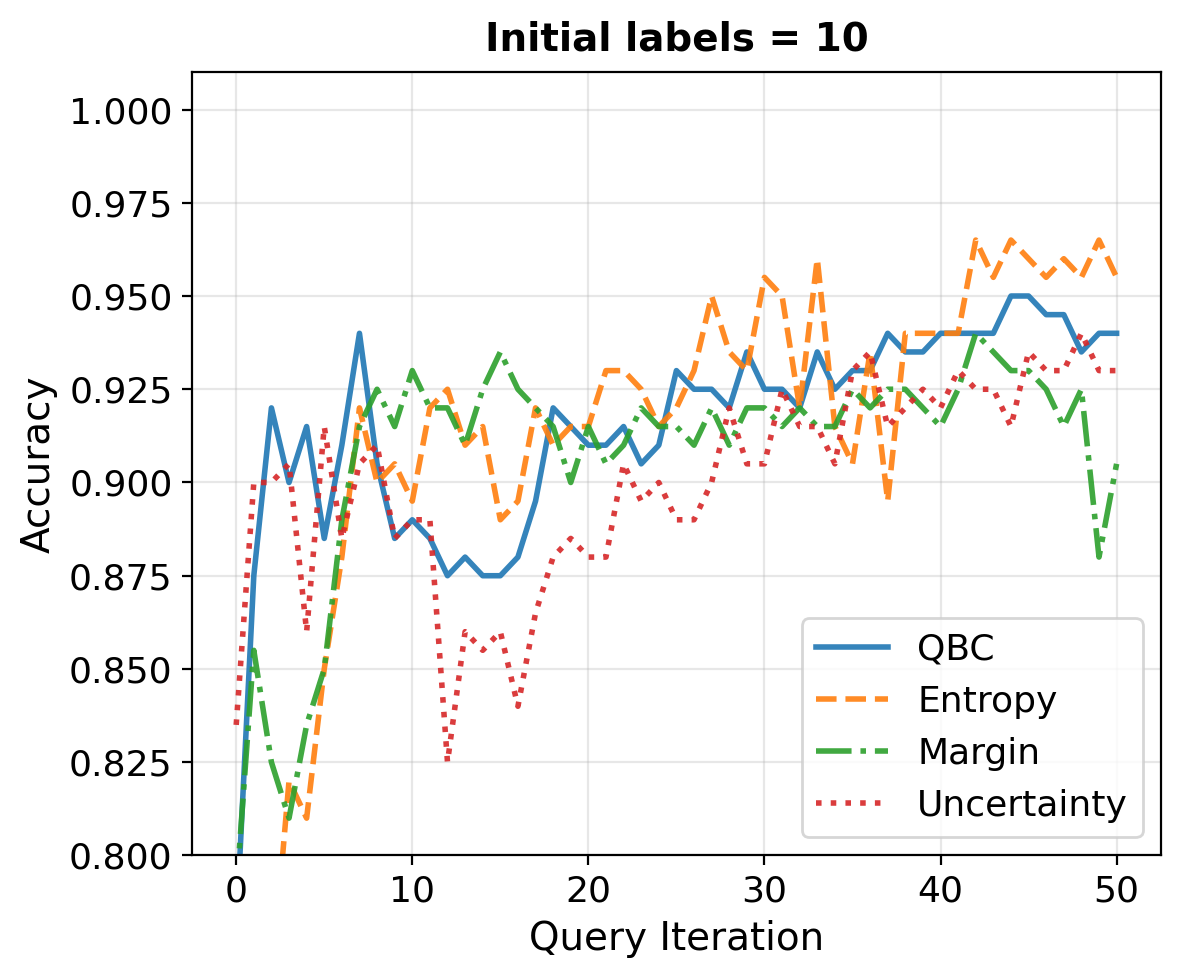}
    \caption{Initial labels=10.}
  \end{subfigure}
  \hfill
  \begin{subfigure}[t]{0.32\textwidth}
    \centering
    \includegraphics[width=\linewidth, trim=0 0 0 0.7cm, clip]{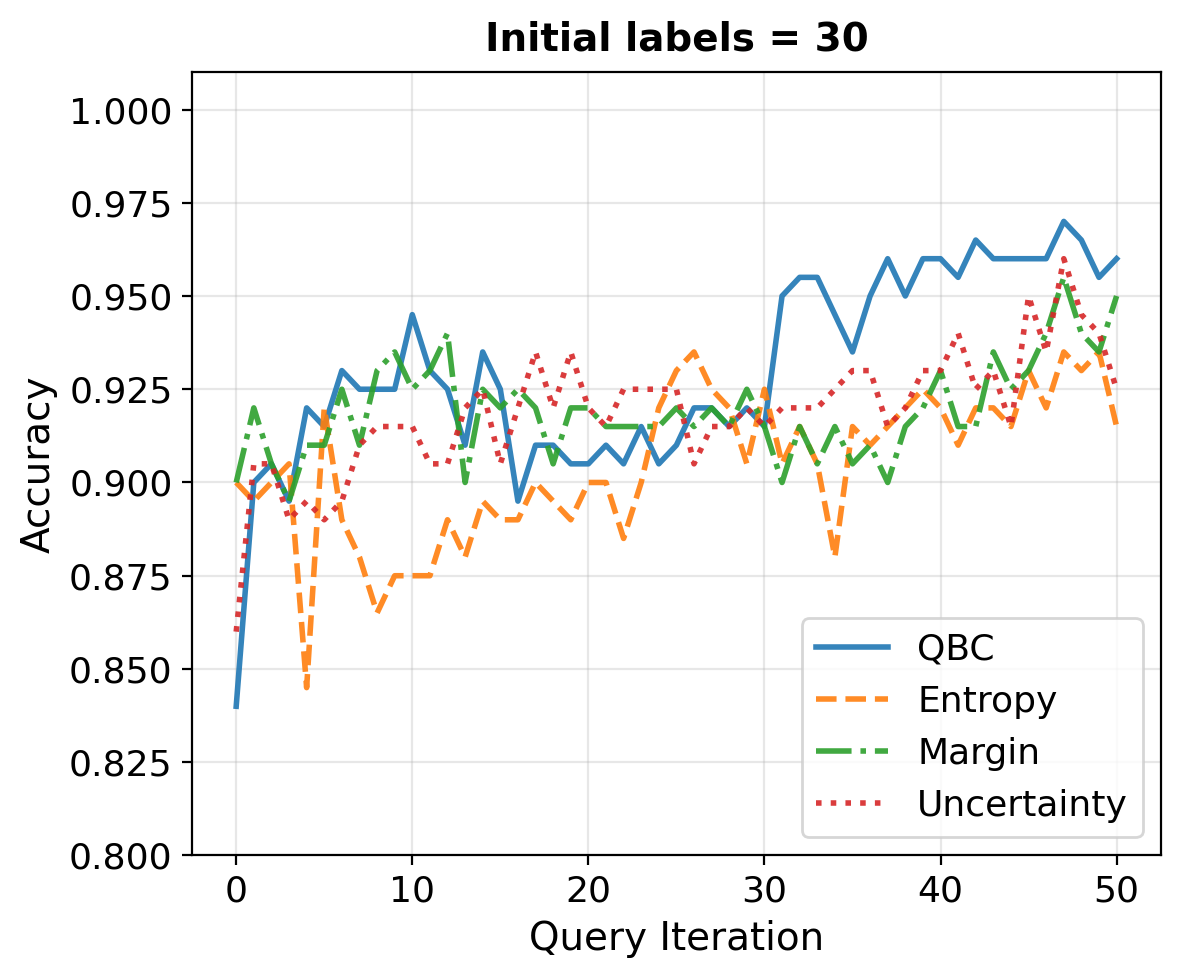}
     \caption{Initial labels=30.}
  \end{subfigure}
  \hfill
  \begin{subfigure}[t]{0.32\textwidth}
    \centering
    \includegraphics[width=\linewidth, trim=0 0 0 0.7cm, clip]{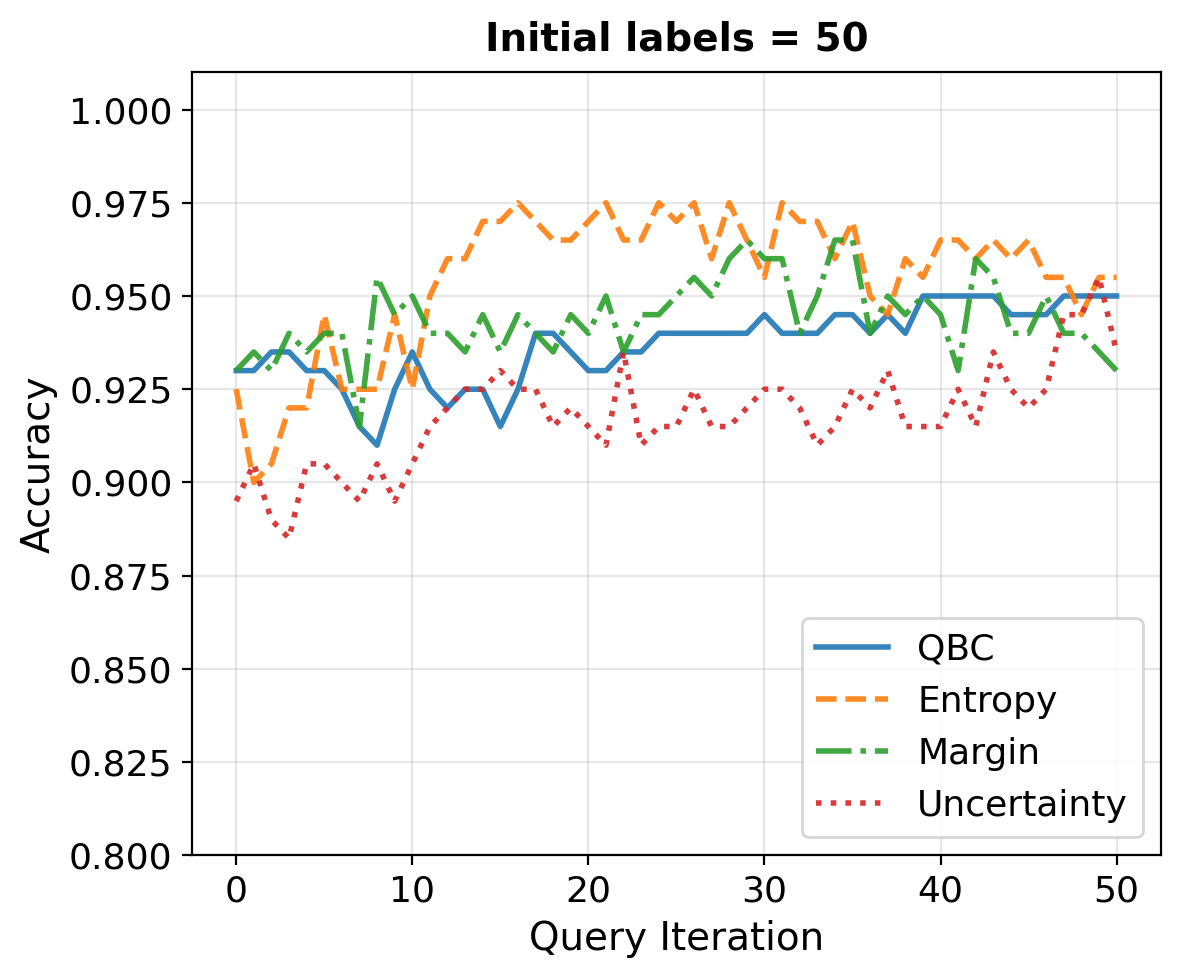}
       \caption{Initial labels=50.}
  \end{subfigure}

  \vspace{3mm}
   \hfill
  \begin{subfigure}[t]{0.32\textwidth}
    \centering
    \includegraphics[width=\linewidth, trim=0 0 0 0.7cm, clip]{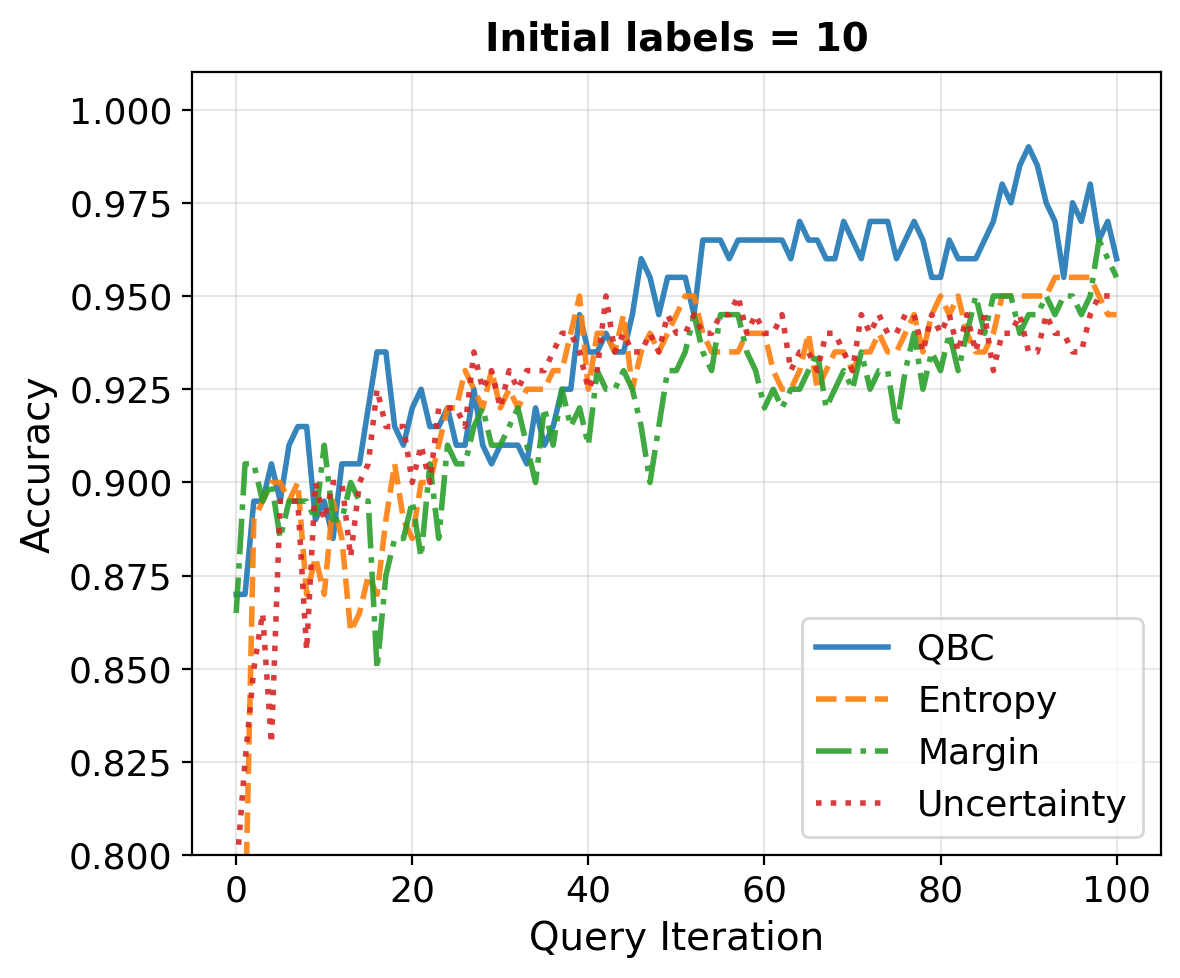}
       \caption{Initial labels=10.}
  \end{subfigure}
   \hfill
  \begin{subfigure}[t]{0.32\textwidth}
    \centering
    \includegraphics[width=\linewidth, trim=0 0 0 0.7cm, clip]{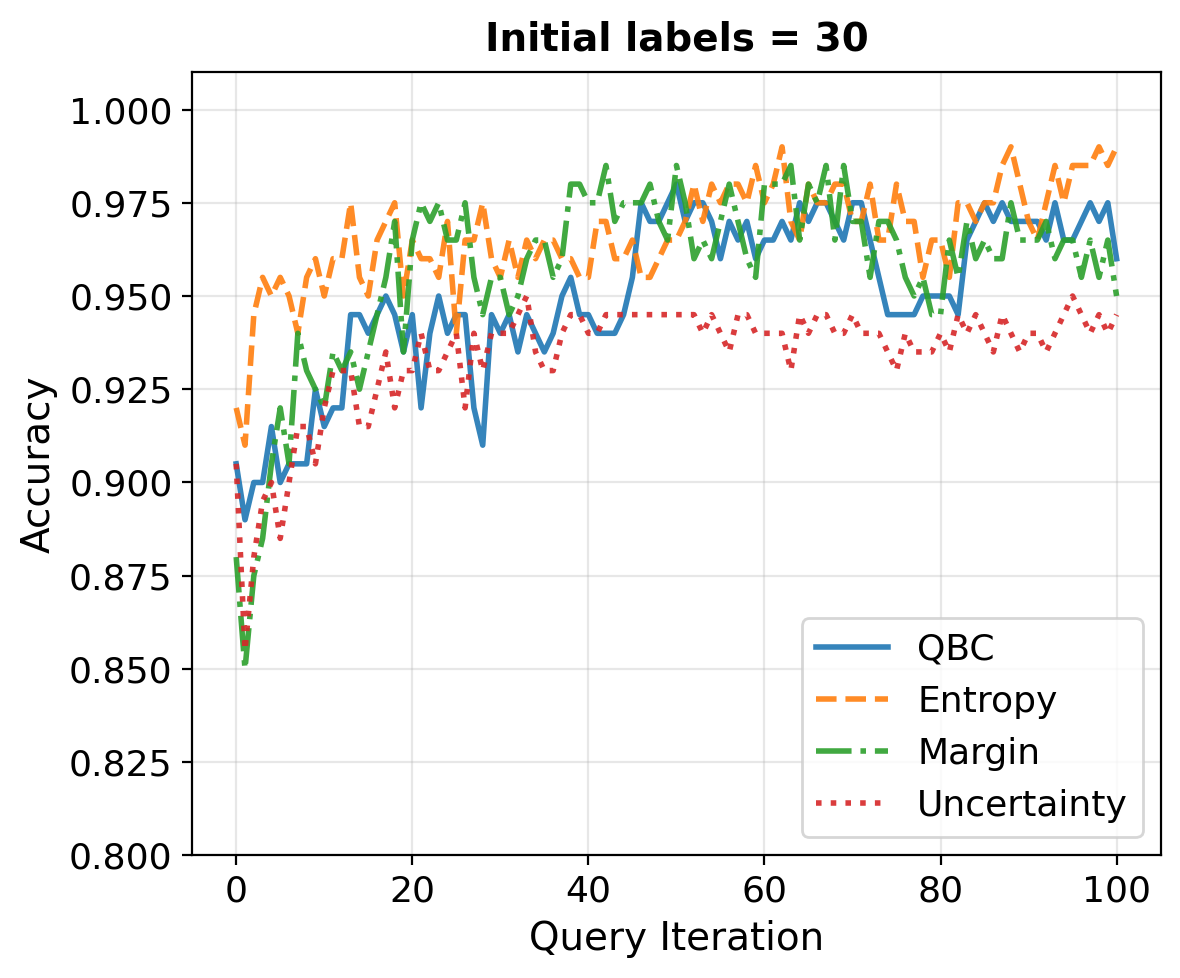}
   \caption{Initial labels=30.}
  \end{subfigure}
   \hfill
  \begin{subfigure}[t]{0.32\textwidth}
    \centering
    \includegraphics[width=\linewidth, trim=0 0 0 0.7cm, clip]{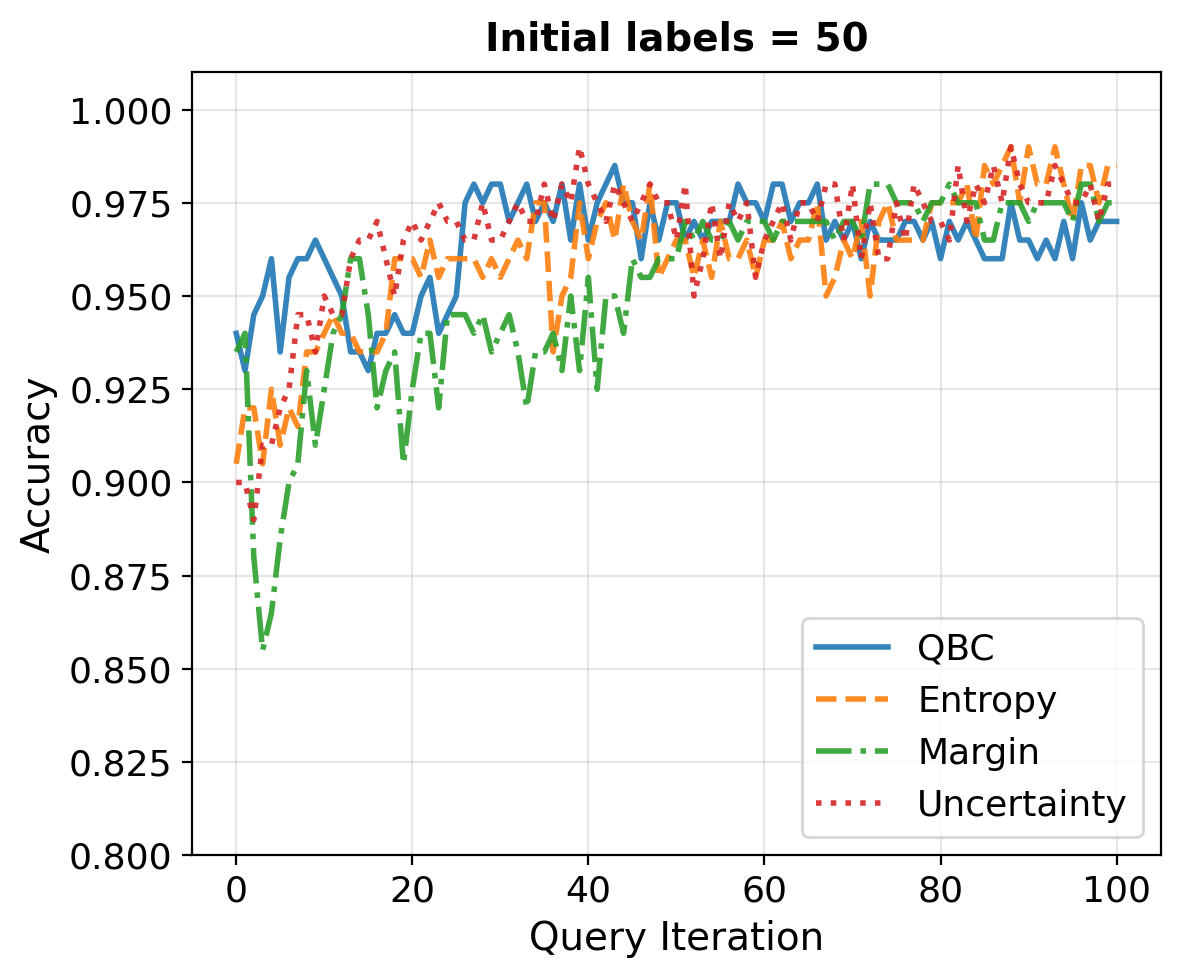}
       \caption{Initial labels=50.}
  \end{subfigure}
\caption{Comparison of active learning query strategies under different query budgets: 50 queries (top) and 100 queries (bottom).}\label{fig:strategy_combined}
\end{figure*}

\vspace{2mm}
\noindent{\bf Performance of Different Query Strategies.} 
To further investigate the efficiency of active learning, we evaluated four query strategies: uncertainty sampling, entropy sampling, margin sampling, and query-by-committee (QBC). Fig.~\ref{fig:strategy_combined} illustrates that all strategies converge as the number of queries increases from 50 to 100.  Notably, even with a small initial labeled set of size 10, entropy and margin sampling achieve up to 94\% test accuracy, while QBC and uncertainty sampling reach approximately 91–93\%, demonstrating the effectiveness of active learning in maximizing information gain from limited labels.

With very few labeled samples, entropy sampling consistently outperforms other strategies, reaching 94\% accuracy with an initial set of 10 samples and 50 queries, increasing to 95\% with 100 queries. Margin sampling shows the largest improvement for slightly larger initial sets; for instance, when the initial labeled set is 30, it reaches 97.5\% accuracy for both 50 and 100 queries. QBC generally matches uncertainty sampling but slightly underperforms with small initial labeled sets, likely due to limited committee diversity early on.

Increasing the number of queries from 50 to 100 provides diminishing yet measurable improvements across all strategies, suggesting that additional samples provide incremental but meaningful information for the classifier. With larger initial labeled sets, strategy performance converges, showing advanced query strategies are most beneficial when labeled data is scarce.

These results highlight that entropy and margin sampling are particularly effective in low-label regimes, efficiently guiding the model toward high accuracy with minimal labeling effort. QBC can be a viable alternative but may require larger initial labeled sets to achieve comparable performance. 


\begin{wraptable}{r}{0.6 \linewidth}
    \centering
    \vspace{-25pt}
    \caption{Accuracy comparison of QBC, Entropy, Margin, and Uncertainty strategies.}
    \label{tab:qbc_compare}
    \vspace{-5pt}
    \resizebox{\linewidth}{!}{
    \begin{tabular}{c|c|c|c|c|c}
    \toprule
    \textbf{Query} & \textbf{Init.} & \textbf{QBC} & \textbf{Entropy} & \textbf{Margin} & \textbf{Uncert.} \\
    \midrule
    \multirow{3}{*}{50} 
        & 10  & 91 & 94 & 92.5 & 93.5 \\
        & 30  & 91.5 & 94 & 97.5 & 94 \\
        & 50  & 96.5 & 96.5 & 97 & 96.5 \\
    \midrule
    \multirow{3}{*}{100} 
        & 10  & 97.5 & 95 & 95.5 & 96 \\
        & 30  & 96.5 & 99 & 97.5 & 98 \\
        & 50  & 96.5 & 95 & 94.5 & 96 \\
    \bottomrule
    \end{tabular}}
    \vspace{-20pt}
\end{wraptable}

\vspace{2mm}
\noindent{\bf Overall Comparison Across Strategies.} Table~1 summarizes the performance of all evaluated active learning strategies, including QBC, uncertainty sampling, entropy sampling, and margin sampling, across varying initial labeled set sizes and query counts.  

For smaller labeled sets, entropy and margin sampling consistently achieve the highest accuracy, highlighting their ability to select the most informative samples in low-data regimes. Uncertainty sampling and QBC also perform well but show slightly higher variability with very small initial labeled sets, reflecting sensitivity to limited data.

As both the initial labeled set size and the number of queries increase, the variance in accuracy across all strategies decreases. In these higher-data scenarios, the differences between strategies become negligible, indicating that all approaches converge toward similar performance when sufficient labeled information is available. 



\begin{wrapfigure}{r}{0.55\linewidth}
    \centering
     \vspace{-4pt}
    \includegraphics[width=\linewidth]{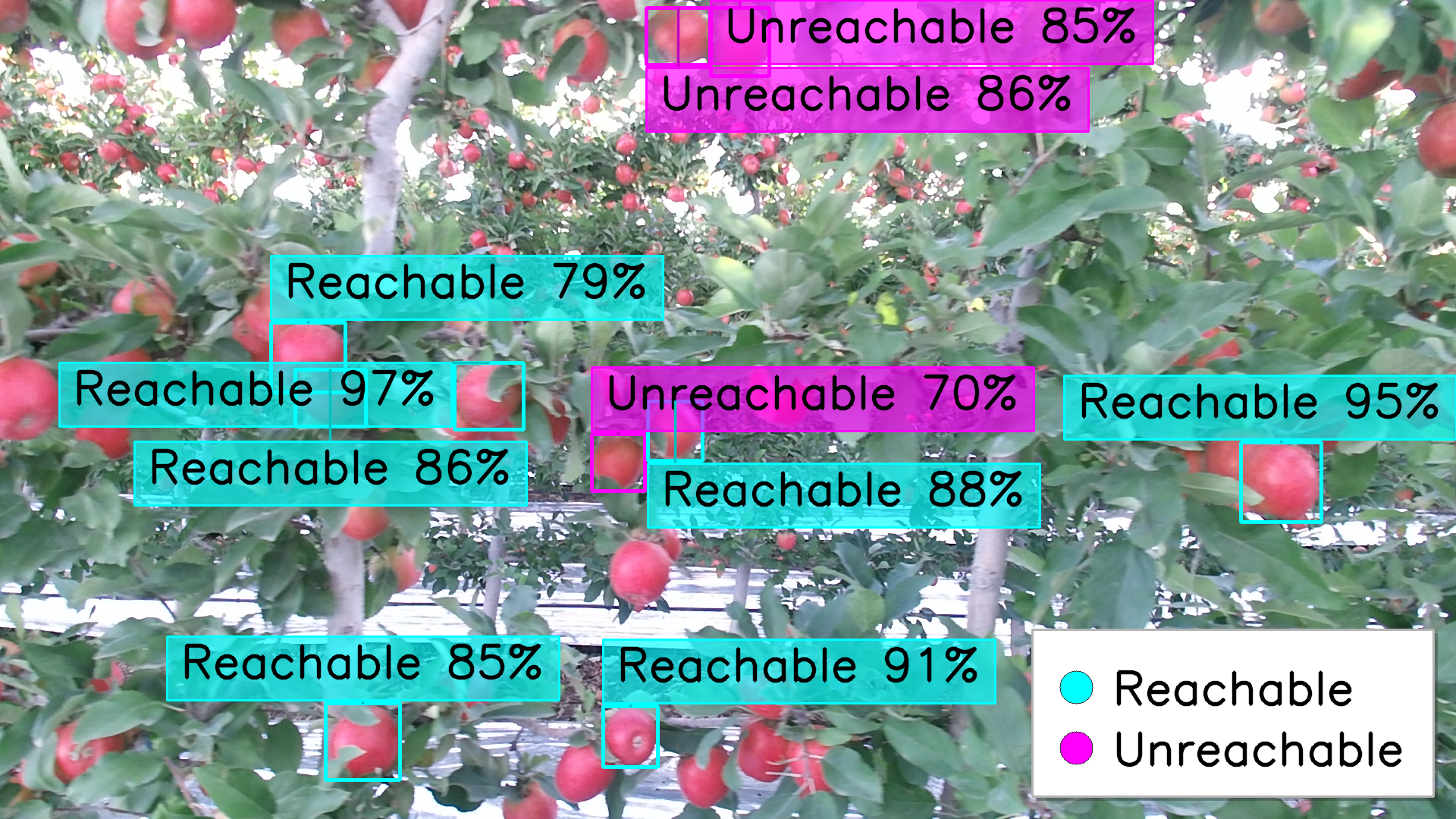}
    \vspace{-10pt}
\caption{Qualitative results of the reachability classifier on a real orchard image. Cyan boxes indicate reachable apples; purple boxes indicate unreachable ones. Confidence scores are shown for each detection.}
    \label{proof} \vspace{-20pt}
\end{wrapfigure}


\vspace{2mm}
\noindent{\bf Qualitative Evaluation.} \fref{proof} shows the qualitative output of the framework on a representative orchard image. Apples in the central region are correctly classified as reachable (cyan), while those near the image periphery, with larger lateral offsets from the arm origin, are classified as unreachable (purple). The predicted confidence scores exceed 0.70 for all detections, indicating reliable confidence. Overall, the visualization shows that the framework captures reachability and separates feasible from infeasible targets in cluttered scenes.

\vspace{-3mm}
\section{Conclusion}


This paper presented a learning-based reachability estimation framework that combines RGB-D perception in real orchard environments with active learning to efficiently predict fruit reachability for robotic harvesting. Experimental results showed that active learning outperformed the random sampling baseline. Further analysis demonstrated that entropy- and margin-based query strategies were more effective than QBC and uncertainty sampling in the low-label regime, while all methods gradually converged as more labeled data became available.

\vspace{-2mm}
\bibliographystyle{splncs04}
\vspace{-1mm}
\bibliography{references}

\end{document}